\documentclass[10pt,twocolumn,letterpaper]{article}

\usepackage[pagenumbers]{wacv} 

\usepackage{graphicx}
\usepackage{amsmath}
\usepackage{amssymb}
\usepackage{booktabs}
\usepackage{ragged2e} 
\usepackage{booktabs,makecell, multirow, tabularx}
\usepackage{footmisc}
\usepackage[pagebackref,breaklinks,colorlinks]{hyperref}

\usepackage[capitalize]{cleveref}
\crefname{section}{Sec.}{Secs.}
\Crefname{section}{Section}{Sections}
\Crefname{table}{Table}{Tables}
\crefname{table}{Tab.}{Tabs.}

\def\wacvPaperID{10} 
\def\confName{WACV Workshop}
\def\confYear{2024}

\begin{document}

\title{A Safer Vision-based Autonomous Planning System for Quadrotor UAVs with Dynamic Obstacle Trajectory Prediction and Its Application with LLMs}

\author{
	Jiageng Zhong$^{\dag,1}$, Ming Li$^{*,\dag,1,2}$, Yinliang Chen$^{\dag 1}$, Zihang Wei$^{3}$, Fan Yang$^{4}$, Haoran Shen$^{1}$\\
	$^{1}$Wuhan University, $^{2}$ETH Zürich, $^{3}$WFLS, $^{4}$HUST}



\maketitle

\begin{abstract}
For intelligent quadcopter UAVs, a robust and reliable autonomous planning system is crucial. Most current trajectory planning methods for UAVs are suitable for static environments but struggle to handle dynamic obstacles, which can pose challenges and even dangers to flight. To address this issue, this paper proposes a vision-based planning system that combines tracking and trajectory prediction of dynamic obstacles to achieve efficient and reliable autonomous flight. We use a lightweight object detection algorithm to identify dynamic obstacles and then use Kalman Filtering to track and estimate their motion states. During the planning phase, we not only consider static obstacles but also account for the potential movements of dynamic obstacles. For trajectory generation, we use a B-spline-based trajectory search algorithm, which is further optimized with various constraints to enhance safety and alignment with the UAV's motion characteristics. We conduct experiments in both simulation and real-world environments, and the results indicate that our approach can successfully detect and avoid obstacles in dynamic environments in real-time, offering greater reliability compared to existing approaches. Furthermore, with the advancements in Natural Language Processing (NLP) technology demonstrating exceptional zero-shot generalization capabilities, more user-friendly human-machine interactions have become feasible, and this study also explores the integration of autonomous planning systems with Large Language Models (LLMs).
\end{abstract}

\section{Introduction}

\renewcommand{\thefootnote}{}
\footnotetext{$^{\dag}$ Equal technical contribution}
\footnotetext{$^{*}$ Corresponding author, E-mail address: mingli39@ethz.ch}

In recent years, small-scale quadrotor Unmanned Aerial Vehicles (UAVs) have experienced rapid development due to their compact size, high agility, and strong flexibility. They have been widely employed in various civilian applications \cite{lu2023comprehensive51,zhou2020ego22,lin2020robust05}, such as industrial inspection, cinematography, and mining operations. These application scenarios typically feature limited space, unknown, variable structures, and the presence of dynamic obstacles, which pose challenges and risks to UAV flight. Therefore, a reliable real-time onboard planning algorithm is a key prerequisite for enabling UAVs to navigate safely, avoid obstacles, and ultimately achieve autonomous flight capabilities in such environments.

The autonomous navigation of UAVs using onboard sensors has been extensively researched and successfully validated in static environments \cite{zhou2020ego22,gao2019flying53,tordesillas2019faster54,falanga2019fast55,chen2021computationally56}. However, in dynamic environments, the presence of dynamic obstacles can pose significant difficulties for UAV flight. Since the motion states of these objects are typically unknown, if the UAV treats them as static, it may not have sufficient time to replan its trajectory as it approaches them, potentially resulting in collisions. Thus, unlike obstacle avoidance in static environments, it is imperative to take into account the motion states of dynamic obstacles in environments. However, dealing with dynamic obstacles proves to be a challenging task, especially the accurate perception of such objects during flight, requiring robust and efficient algorithms. Furthermore, considering the limited computational capabilities of onboard computers, which are responsible for environmental perception, navigation, positioning, and interaction functionalities. So, evidently, it is hard to satisfy the efficiency and accuracy requirements at the same time. Existing obstacle avoidance systems \cite{chen2021active57,chen2023risk33,wang2021autonomous02,zhu2019chance04,lin2020robust05} in the dynamic environments of quadrotors typically rely on the detection and tracking of moving objects (DATMO) \cite{wang2007simultaneous72} to acquire the motion states of dynamic objects and process them in various ways for subsequent planning. However, these methods still have some shortcomings in terms of dynamic obstacle detection or obstacle avoidance capabilities.

In this paper, we propose a vision-based autonomous planning system for quadrotor UAVs that can predict the future trajectories of dynamic obstacles and generate safer trajectories. First, a very efficient object detection network, NanoDet \cite{nanodet37}, is first applied to detect dynamic obstacles, followed by Kalman Filtering (KF) to track and estimate their motion states. Then, the static and the dynamic obstacles are both considered in our planning process. We utilize a B-spline-based search algorithm, optimized with several constraints to enhance safety and smoothness. The effectiveness of our system is verified through various experiments in simulation and real-world environments. Unlike planners that rely on traditional methods for detecting dynamic obstacles \cite{chen2023risk33,lin2020robust05,wang2021autonomous02}, the detector used in our method is based on deep learning. Consequently, it not only provides information about the sizes and locations of the objects but also offers rich semantic data, enabling broader potential applications in the future.

To make drones more user-friendly, further support from human-machine interaction technology is also essential. Natural Language Processing (NLP) has long been recognized as an important way of human-robot interaction, and the recent rapid advancement has given rise to the development of Large Language Models (LLMs), which have demonstrated remarkable performance in different applications \cite{vemprala2023chatgpt46}. Several LLMs have already achieved significant accomplishments, such as BERT \cite{devlin2018bert60}, GPT-3 \cite{brown2020language61}, GPT-4 \cite{chatgpt63} and Codex \cite{chen2021evaluating62}. The most influential among them is ChatGPT \cite{chatgpt59}, which is a pre-trained generative text model that underwent fine-tuning through human feedback. Recent studies also demonstrate LLMs can be applied to generalize robotics domains \cite{vemprala2023chatgpt46,palnitkar2023chatsim47,liu2023llm48}. With the support of LLMs, autonomous UAVs undoubtedly gain enhanced versatility. In this paper, we explore how our proposed autonomous planning system integrates with LLMs to assist users in controlling UAVs better, with the hope of providing valuable insights for subsequent research in this domain.

\section{Related work}

In environments with dynamic obstacles, a robust and reliable autonomous flight system typically requires two essential components: one is the dynamic obstacle perception, and the other is the trajectory planner designed for obstacle avoidance in dynamic environments. 

For obstacle detection and tracking, the most commonly used sensors are RGB cameras \cite{xie2017towards65,chen2019monocular66} or depth cameras \cite{oleynikova2015reactive41,xu2023real11,lin2020robust05}. Depth cameras, in particular, are favored for small-sized UAV navigation since they provide both images and point cloud data. Some methods \cite{oleynikova2015reactive41,xu2023real11,lin2020robust05}utilize U-depth maps extracted from depth maps for obstacle detection. Among them, \cite{lin2020robust05} has achieved impressive results in obstacle avoidance in dynamic environments. While these methods exhibit high computational efficiency, their performance stability is contingent on parameter settings, and they do not possess the capability to recognize objects. \cite{mori2013first67} applies a 2D feature-based method to achieve obstacle avoidance. On the other hand, \cite{tai2016deep68,eppenberger2020leveraging15} utilizes machine learning techniques to enhance performance and acquire semantic information about objects, albeit at a higher computational cost. Additionally, there are methods based on point cloud data \cite{wang2021autonomous02,chen2023risk33,chen2021active57,min2021kernel69}, which handle dynamic obstacles by building a dynamic map for both static and dynamic obstacles. \cite{wang2021autonomous02} integrates dynamic obstacle detection results into the static map generated from the depth point cloud, while \cite{chen2021active57} utilizes the voxel map to identify dynamic voxels and estimate their velocities. \cite{chen2023risk33} builds a dual-structure particle-based dynamic occupancy map to concurrently depict the static obstacles and dynamic obstacles. To achieve efficient and robust dynamic obstacle perception, we apply a very efficient object detection neural network, and the semantic information obtained is also very useful.

For obstacle avoidance, various strategies have been explored in static environments. The most common pipeline is to detect obstacles and represent them in the occupancy map \cite{usenko2017real23}, and then generate a safe flight trajectory using optimization-based methods \cite{zhou2020ego22,liu2017planning25,zhou2019robust01,chen2016online71} or sampling-based methods \cite{dharmadhikari2020motion70,chen2021computationally56}. Among these, EGO-Planner \cite{zhou2020ego22} stands as one of the most representative methods, distinguished by its robustness and efficiency. However, in dynamic environments, obstacle avoidance becomes more demanding, as it entails the prediction of future states for dynamic obstacles \cite{chen2023risk33}. Currently, obstacle avoidance systems for quadrotor UAVs in dynamic environments typically apply DATMO \cite{wang2007simultaneous72} for dynamic obstacle perception, followed by the application of modified planning algorithms to generate suitable flight trajectories. The ASAA system \cite{chen2021active57} uses YOLOv3 \cite{redmon2018yolov3_73} for dynamic obstacle detection, subsequently utilizing SORT \cite{bewley2016simple74} and active vision for object tracking. It then performs trajectory planning based on real-time sampling and uncertainty-aware collision checking, ultimately enabling the UAV to avoid collisions with slow-moving objects. Similarly, \cite{wang2021autonomous02} also applies a detection and tracking method, and further optimizes trajectories by considering the predicted position of the dynamic obstacles, thereby ensuring flight safety. In order to address the uncertainty in trajectory prediction, \cite{lin2020robust05} develops a chance-constrained model predictive controller to ensure that the collision probability between the UAV and each moving object remains acceptably low, thus achieving robust obstacle avoidance. \cite{chen2023risk33} achieves efficient trajectory planning through sampling motion primitives in the state space and forming risk with the cardinality expectation in a risk-checking corridor. Based on robust mobile object detection, our method features trajectory prediction and reliable trajectory optimization under several constraints, possessing good generality and user-friendliness.

\section{Method}

\subsection{System framework}

The whole vision-based autonomous planning system can be divided into two components, as shown in Figure 1. The first component, the perception module, is primarily dedicated to the construction of local maps and the identification, tracking, and prediction of dynamic obstacles. On the other hand, the second component, the trajectory generation module, serves the purpose of conducting an initial trajectory search and subsequent trajectory optimization. The input of the system consists of RGB images, depth maps, and the UAV pose. RGB images and depth maps can be directly acquired through an RGB-D camera such as the Realsense D435i, while precise UAV pose information can be obtained using Simultaneous Localization and Mapping (SLAM) techniques, such as VINS-Fusion \cite{qin2019general52}. These data sources are synchronized using timestamps and then fed into the perceptual module. The output is a collision-free and dynamically feasible trajectory leading to the target point.


\begin{figure}
	\centering
	\includegraphics[scale=0.0665]{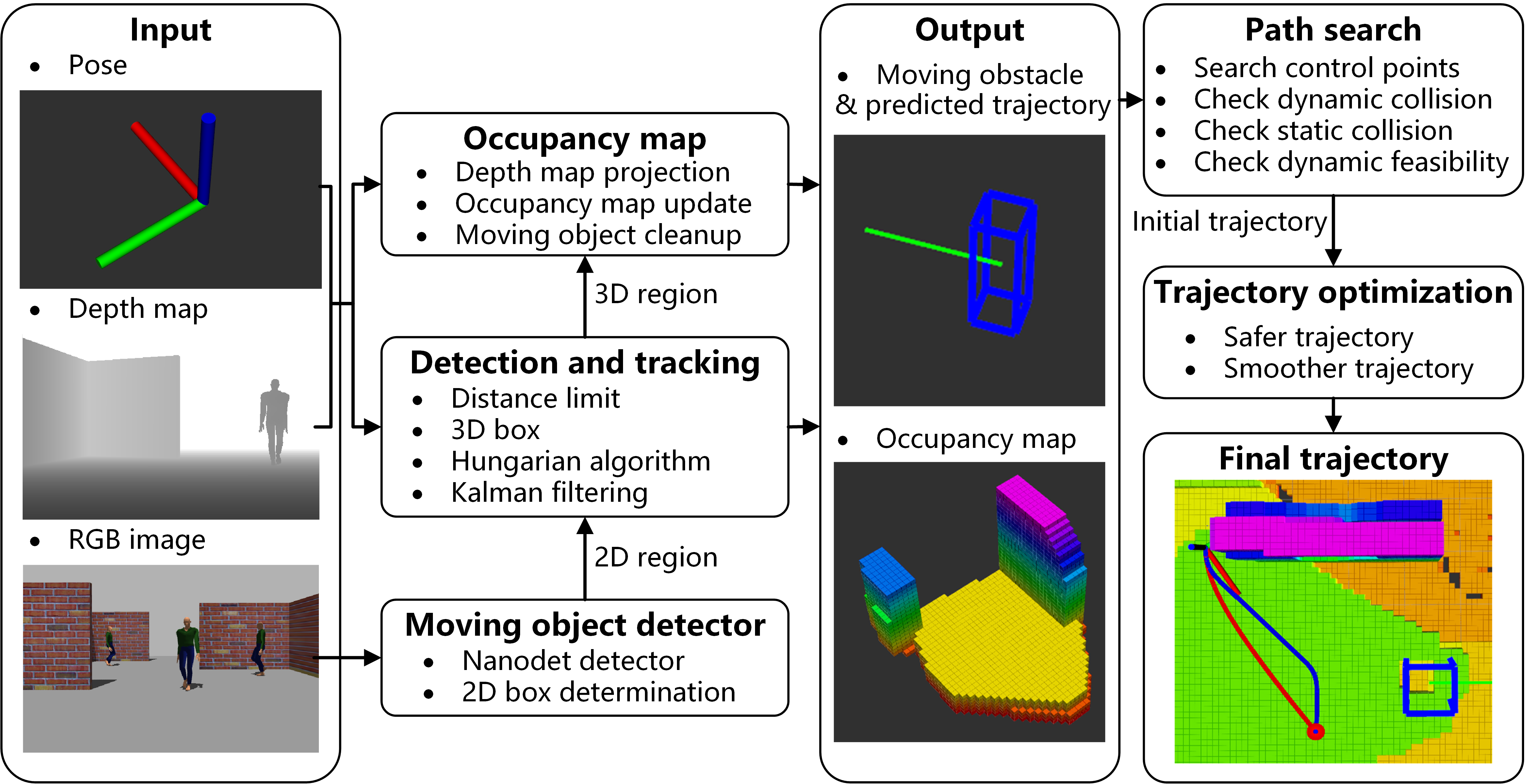}
	\caption{The framework of our autonomous planning system.}
	\label{fig1}
\end{figure}

In the first stage, the UAV is tasked with the comprehensive perception of its surroundings, encompassing both the static environment and dynamic obstacles. RGB images are input into a lightweight object detection neural network, NanoDet \cite{nanodet37}, for the purpose of dynamic obstacle detection, yielding 2D regions corresponding to the detected dynamic obstacles. Subsequently, these 2D regions are combined with depth data and UAV poses to derive their respective 3D regions. Further, the Hungarian algorithm is employed for multi-object tracking, enabling the determination of trajectories for the dynamic obstacles over a time interval. Furthermore, the real-time trajectory prediction of dynamic obstacles is achieved using KF algorithm. Additionally, the depth data is employed for the construction of local occupancy maps representing the static components of the scene. In the second phase, inspired by the work of Tang and et al. \cite{tang2019real28}, we design a trajectory generation module. The trajectory planning process can be considered as the task of identifying a set of B-spline control points. This problem is analogous to a graph optimization problem and can be effectively resolved. Based on occupancy maps, the process initiates with the generation of an initial trajectory employing the A* algorithm. Subsequently, an optimization function is formulated, accounting for constraints related to obstacles, UAV velocity, and acceleration. This function undergoes optimization through a solver, ultimately yielding a final trajectory characterized by improved safety and smoothness.

\subsection{Dynamic obstacle detection and tracking}

\begin{figure}
	\centering
	\includegraphics[scale=0.18]{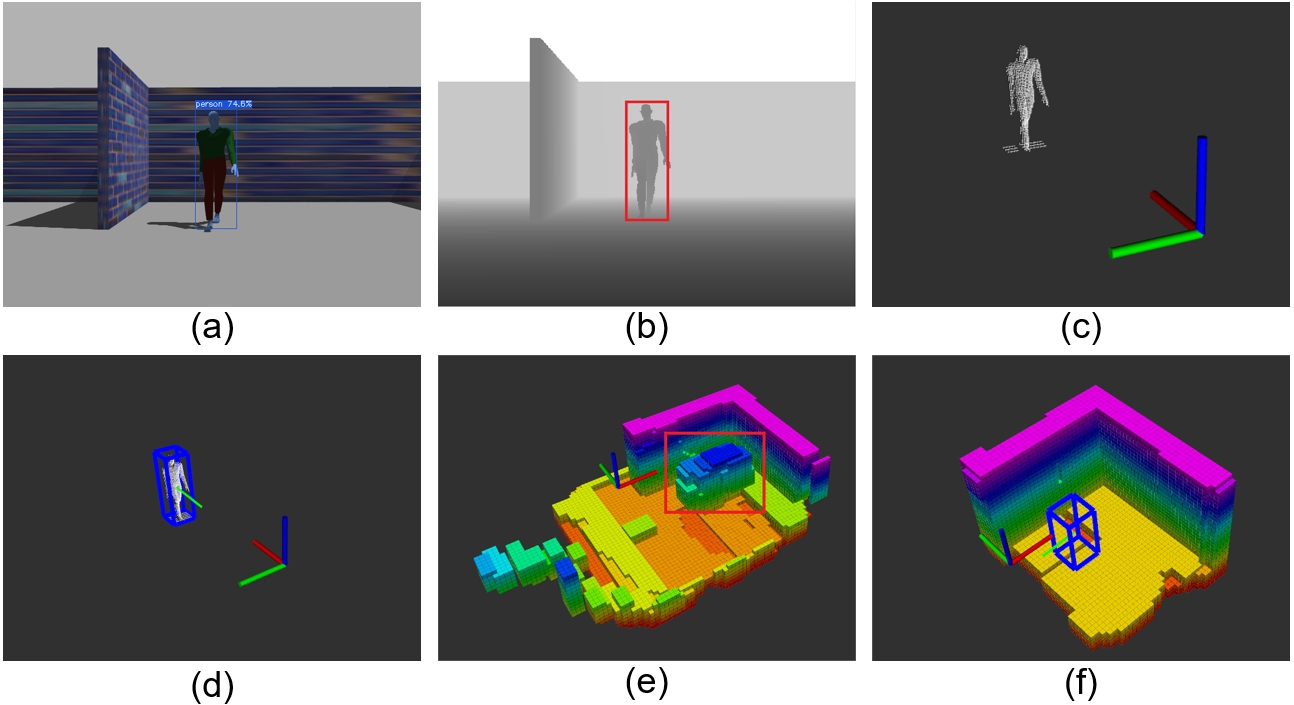}
	\caption{(a) The 2D region of the dynamic obstacle generated by object detection. (b) The corresponding position of the object in the depth map. (c) The depth point cloud of the object. (d) The object tracking and trajectory prediction. (e) Residuals of the dynamic obstacle in the local map. (f) Residual clearance and differentiation of the static scene and dynamic obstacle.}
	\label{fig2}
\end{figure}

To avoid collisions with dynamic obstacles, the first step is to determine the state of these dynamic obstacles, which requires detection and tracking. As depicted in Figure \ref{fig1}, the initial phase involves utilizing captured RGB images to obtain 2D regions of potential moving objects (as shown in Figure \ref{fig2} (a)). Given the limited computing resources of UAV, a lightweight neural network, NanoDet, is employed for this purpose. In comparison to point cloud-based clustering algorithms \cite{oleynikova2015reactive41,jafari2014real42}, this detection method exhibits greater robustness, with reduced sensitivity to algorithm parameters and environmental variations, while simultaneously delivering efficiency and high accuracy. Additionally, the extracted semantic information can also contribute to further human-machine interaction. After acquiring the 2D regions of objects, it becomes feasible to determine their positions in the depth map, as shown in Figure \ref{fig2} (b). Finally, the depth information of obstacles can be extracted with the elimination of background depth values. This procedure enables the extraction of 3D information about the objects, including 3D regions and fine structural details.

Based on the results of object detection, object tracking can be performed. By calculating the pairwise distances between the sets of detected objects in two different time-step images, the Hungarian algorithm is applied to obtain the optimal assignment. This allows the association of information across multiple frames, facilitating object tracking. Furthermore, with the poses of cameras, it becomes feasible to estimate the object's velocity, trajectory, and other state information over a certain time period. The motion status of an object is determined based on its velocity. If its estimated velocity consistently exceeds the predefined velocity threshold $v_d$, it is classified as a moving object; otherwise, it can be considered a component of the static map.

Typical autonomous planning systems utilize a local map for planning, however, without special handling, these systems often incorporate the depth point cloud of detected dynamic obstacles into the map, rather than removing it. However, if the depth point cloud of dynamic obstacles is added to the local map, the UAV may perceive an obstruction in its path, as depicted in Figure \ref{fig2} (c). Therefore, it is essential to exclude dynamic obstacles from the static map and clear their motion histories. In practice, based on the previously detected 3D regions, we can directly remove occupancy voxels at the current positions of dynamic obstacles, effectively eliminating their interference and obtaining a static map for subsequent planning.

Our planning system for avoiding dynamic obstacles is based on predicting their future paths. Predicting the future trajectories of objects has consistently been an interesting subject, with diverse methods \cite{pfeiffer2018data43,wulfmeier2017large44,wang2022stepwise45}. However, given the need to control UAV flight efficiently, the prediction algorithm must be highly efficient. Following the work of Eppenberger \cite{eppenberger2020leveraging15}, we adopt a conservative motion model to estimate the velocities and short-term future trajectories of detected dynamic obstacles. Assuming dynamic obstacles move on a horizontal plane, their velocity can be estimated using Kalman Filter (KF). For a dynamic obstacle, its state vector $ \vec{x}=[x,y,v_x,v_y]$ represents its position and velocity on the horizontal plane. The measurement input $\vec{z}_t$ at time $t$ for KF is the centroid of the object measured in the horizontal plane. The system dynamics and measurement model are defined as: 
\begin{equation}
\vec{x}_{t+1}=A \cdot \vec{x}_{t} + Q,
\end{equation}
\begin{equation}
\vec{z}_{t} = H \cdot \vec{x}_{t} + R,
\end{equation}
where $A$ is the state-transition model, $H$ is the observation model, and $Q$ and $R$ model the system noise and measurement noise, respectively. This allows for real-time short-term prediction of dynamic obstacle trajectories. It should be noted that if a significant deviation between the measured target position and the position estimated by KF is detected during tracking, it is considered as a tracking error. In such cases, previous results are cleared, and tracking is reinitiated.


\subsection{Trajectory generation and optimization}

\begin{figure}
	\centering
	\includegraphics[scale=0.14]{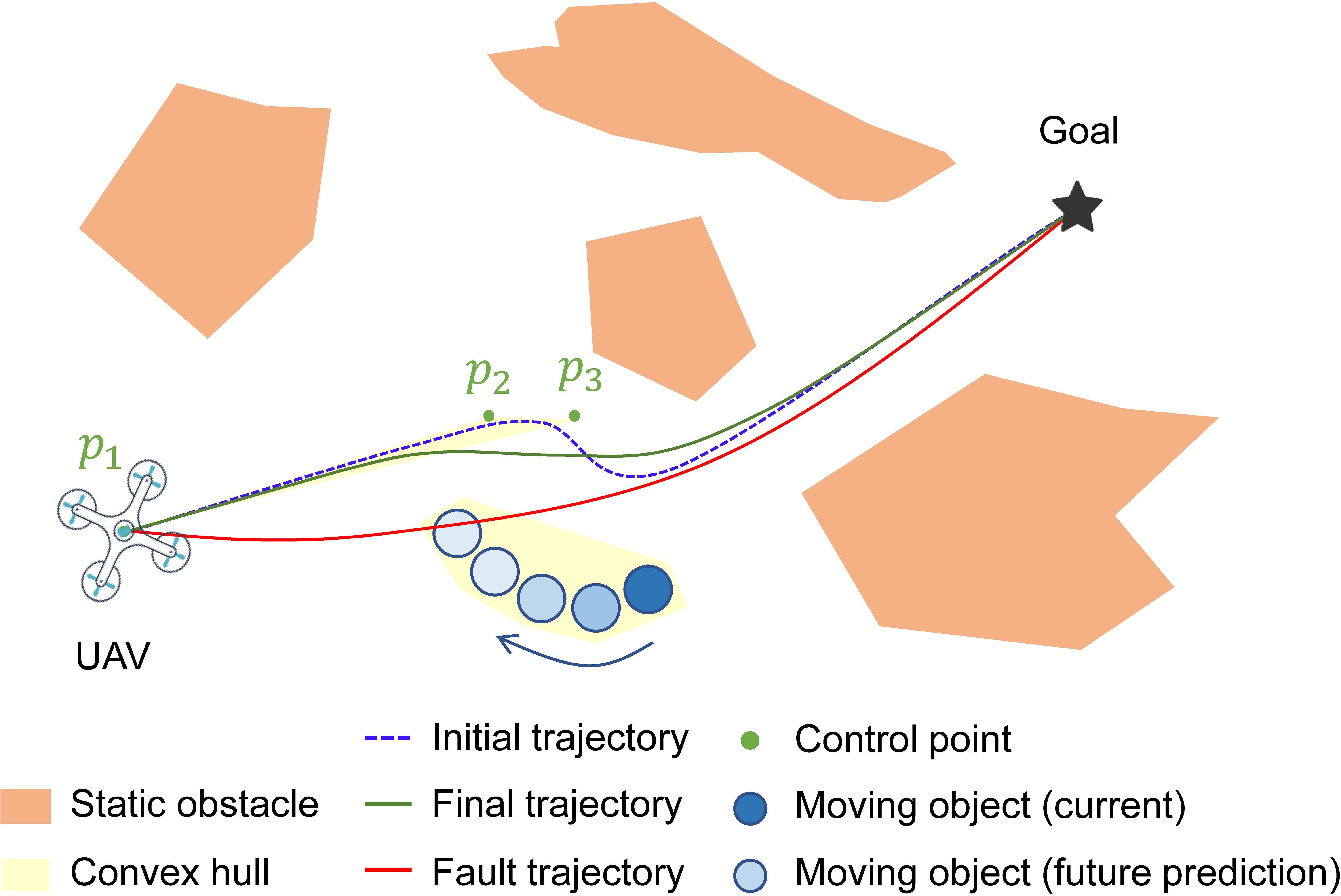}
	\caption{Trajectory generation. This stage takes into account not only the static environment but also dynamic obstacles and their future possible trajectories. }
	\label{fig3}
\end{figure}

In this section, we present a trajectory planning algorithm suitable for dynamic environments. Many existing algorithms either solely consider static environments \cite{zhou2019robust01,zhou2020ego22} or focus exclusively on avoiding dynamic objects \cite{lin2020robust05,tordesillas2021mader06}, making it challenging to meet the condition of collision-free trajectories entirely. Inspired by \cite{tordesillas2021mader06,tang2019real28}, we introduce a trajectory generation method based on B-spline curves and convex hull collision detection, as B-splines can be used to ensure the required smoothness of the trajectory. This approach can be divided into two phases: initial trajectory search and trajectory optimization. It not only ensures collision avoidance in static environments but also takes into account dynamic obstacles and their future motion states. Specifically, we employ collision detection between convex hulls to rapidly assess the safety of trajectories. As illustrated in Figure \ref{fig3}, for instance, the initial trajectory comprises three B-spline control points, $p_0$,$p_1$,$p_2$. During the search process, control points are extended in multiple directions, and the trajectory is enclosed within the convex hull formed by these control points. Additionally, the convex hulls of dynamic obstacles are modeled, requiring collision checking with dynamic and static obstacles for each generated trajectory while preserving dynamic constraints. This process ultimately yields a trajectory to the target point.

In the planning process, the initial trajectory search is essential, as obtaining a reasonable initial trajectory can reduce the optimization time. Inspired by the A* algorithm \cite{hart1968formal49} and B-spline-based Non-uniform Kinodynamic (BNUK) search \cite{tang2019real28}, we apply the A* algorithm to directly search for control points within the free space. The cost function of the algorithm is defined as follows: 
\begin{equation}
F=g+\lambda h,
\end{equation}
where $g$ represents the smoothness of the curve above, $h$ is the heuristic function, and $\lambda$ is the bias, employed to standardize units and to fine-tune the quality of the curve. Here, the heuristic function is defined as the distance to the target point. A smoother curve can lead to less trajectory optimization time and enhance the feasibility of UAV flight control, and the heuristic function aids in guiding the trajectory closer to the target point. To ensure that the B-spline curve passes through the start and target points, we position the initial three control points at the starting location. Likewise, we apply the same treatment to the control points at the target point. Moreover, to guarantee that the obtained trajectories are both smooth and continuous, the latter three control points of the current curve segment are used as the initial control points. This ensures that each trajectory segment is uniformly smooth and seamlessly connected.

\begin{figure}
	\centering
	\includegraphics[scale=0.115]{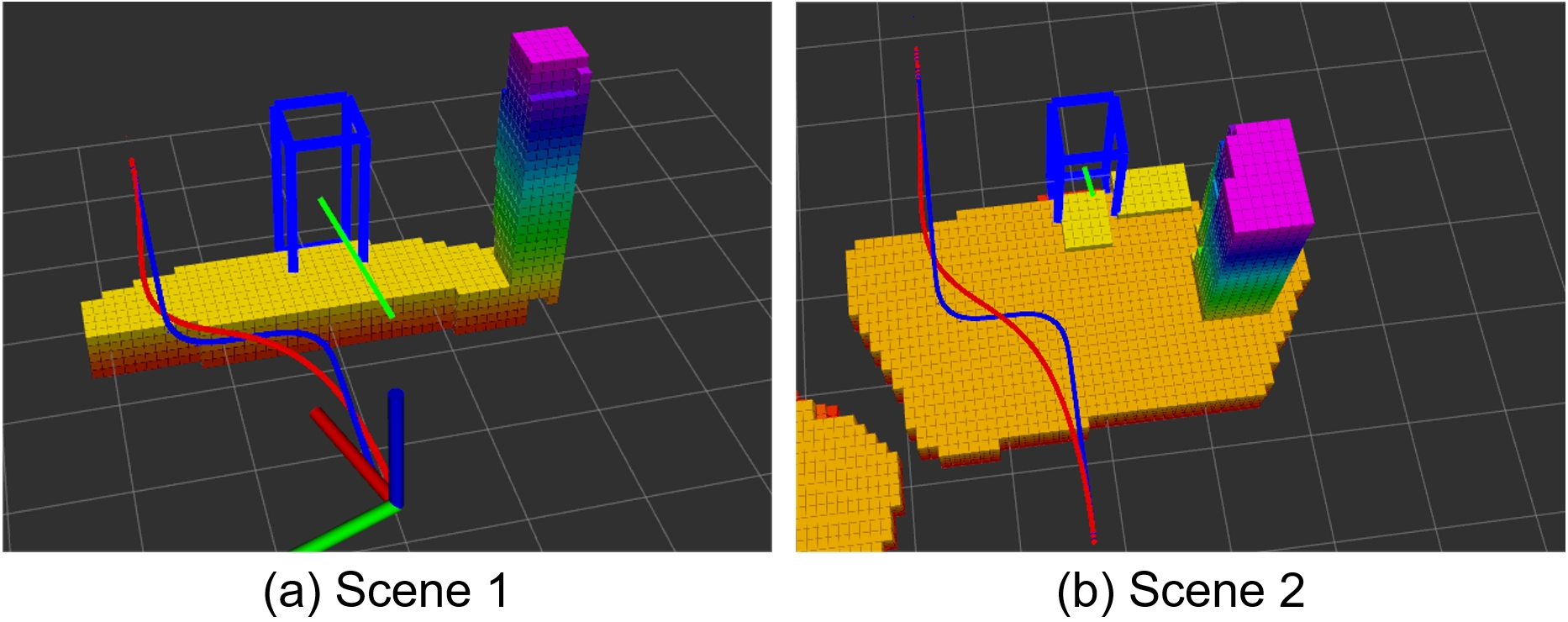}
	\caption{Trajectory Optimization. The blue line represents the trajectory before optimization, the red line represents the trajectory after our optimization, the blue rectangular wireframe represents the bounding box of the dynamic obstacle, and the green line denotes its predicted future trajectory. After optimization, the trajectory becomes smoother.}
	\label{fig4}
\end{figure}

Although the initial trajectory ensures collision avoidance, it may still exhibit issues such as proximity to obstacles or excessive changes in direction, as shown in Figure \ref{fig4}. Therefore, further optimization is required to achieve a trajectory that is not only safer but also better suited for UAV flight. Following Usenko et al. \cite{usenko2017real23}, we represent this problem as an optimization of the following cost function:
\begin{equation}
E_{total}=E_{collision}+E_{smoothness},
\end{equation}
where $E_{collision}$ is a collision cost function and $E_{smoothness}$ is the cost of the integral over the squared derivatives (acceleration, jerk, snap). The former ensures that the UAV will not collide with the static environment or moving targets, while the latter contributes to a smoother trajectory, aiding in flight control. Furthermore, assuming the plane that separates the convex hull of control points and the obstacles is defined as $n^T x+d=0$, where $n$ and $d$ are the parameters of this plane, the optimization must also satisfy the following conditions:
\begin{equation}
	\left \{
	\begin{aligned}
		n^T O +d \geq 0 \\
		n^T Q +d \leq 0 \\
		abs(v) \leq v_{max} \\
		abs(a) \leq a_{max} 
	\end{aligned}
	\right.
	,
\end{equation}
where $O$ represents the vertices of the obstacle's convex hull, and $Q$ represents the control points of the trajectory, $abs(\cdot)$ represents the element-wise absolute value, and $v$ and $a$ represent velocity and acceleration of the UAV, respectively. In this study, this problem is solved using the augmented Lagrangian method \cite{conn1991globally38,birgin2008improving39}, and with the low-storage BFGS algorithm \cite{liu1989limited40} for local optimization. The interface used for these algorithms is NLopt \cite{NLopt36}.

\section{Experiments}

We test our method in both simulation and real-world environments. All the methods are implemented in C++ and executed using the robot operating system (ROS) \cite{quigley2009ros50} in Linux. Sections 4.1 and 4.2 present experiments in the simulation environments. The Gazebo simulation environment with the PX4 firmware is utilized. Section 4.1 covers experiments about dynamic obstacle detection and tracking, and Section 4.2 focuses on assessing the dynamic obstacle avoidance capability. In Section 4.3, we conduct relevant experiments by implementing our method on a real UAV. 

\subsection{Dynamic obstacle detection and tracking}

\begin{figure}
	\centering
	\includegraphics[scale=0.14]{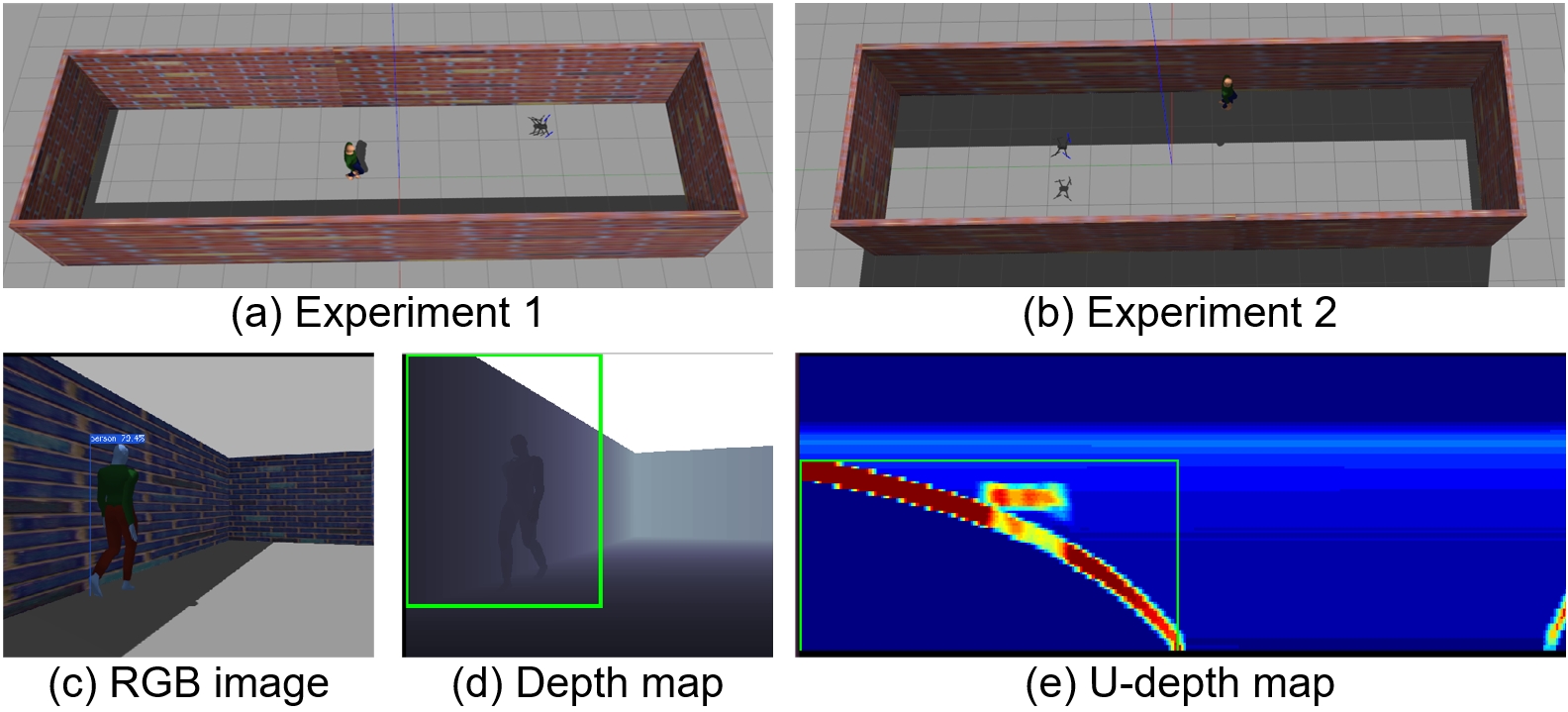}
	\caption{The simulation experiments and object detection examples. (a) and (b) display the overhead views and reconstructed maps of two experiments in this simulation environment. (c) shows the detection results on the RGB image, (d) presents the detection results on the depth map, and (e) is the U-depth map corresponding to the depth map in (d).}
	\label{fig5}
\end{figure}

For quantitative evaluation, we design a simulation environment, as depicted in Figure \ref{fig5} (a). This setting represents a corridor measuring 15 meters in length and 4 meters in width, where a pedestrian moves back and forth. Initially, the UAV and the pedestrian approach each other. In this experiment, our method is compared with three current detectors, denoted as Method \uppercase\expandafter{\romannumeral 1} \cite{wang2021autonomous02}, Method \uppercase\expandafter{\romannumeral 2} \cite{lin2020robust05} and Method \uppercase\expandafter{\romannumeral 3} \cite{xu2023real11}. During the flight, the UAV estimates the positions and velocities of the pedestrian, and these measurements are compared with the ground truth values provided by the simulation environment. The comparison is performed by calculating the Root Mean Square Error (RMSE) for both position and velocity. The results are shown in Table \ref{tab1}. In terms of position estimation, there is little difference in accuracy among the four methods. However, concerning velocity estimation, our method exhibits higher accuracy, surpassing Method \uppercase\expandafter{\romannumeral 1} and Method \uppercase\expandafter{\romannumeral 2}, albeit slightly inferior to Method III.

\begin{table}
	\captionsetup{skip=3pt}
	\caption{The RMSE of the estimated positions and velocities of the pedestrian.}
	\label{tab1}
	\begin{tabular}{@{}lcc@{}}
		\toprule
		Method & Position error (m) & Velocity error (m/s) \\
		\midrule
		Method \uppercase\expandafter{\romannumeral 1} \cite{wang2021autonomous02}	&	0.11 &	0.19 \\
		Method \uppercase\expandafter{\romannumeral 2} \cite{lin2020robust05}	&	0.14 &	0.36 \\
		Method \uppercase\expandafter{\romannumeral 3} \cite{xu2023real11}	&	0.11&	0.08 \\
		Our method &	0.14 &	0.16 \\
		\bottomrule
	\end{tabular}
\end{table}

\begin{figure}
	\centering
	\includegraphics[scale=0.129]{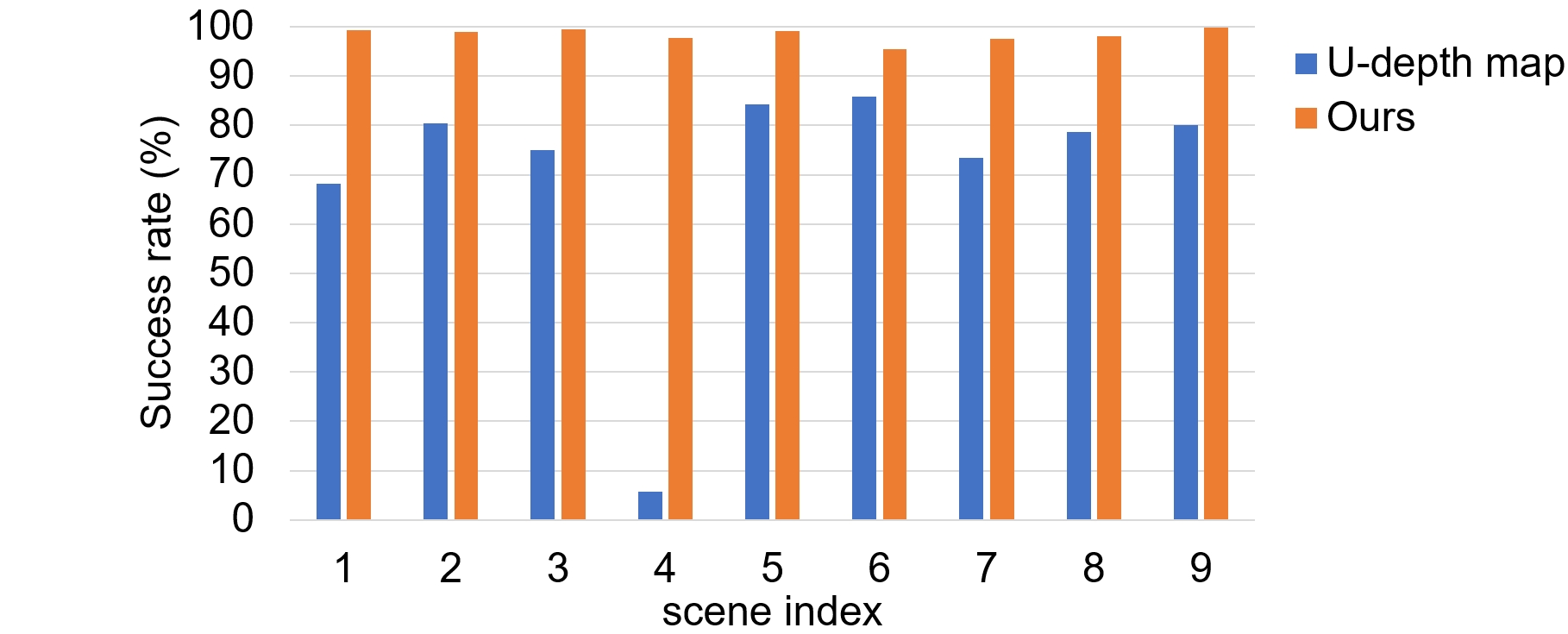}
	\caption{ The success rate of moving target detection.}
	\label{fig6}
\end{figure}

Method \uppercase\expandafter{\romannumeral 2} and Method \uppercase\expandafter{\romannumeral 3} are based on U-depth map, and the U-depth map is derived from the depth map. It's worth noting that although Method \uppercase\expandafter{\romannumeral 3} achieves higher accuracy, it may fail in certain scenarios. For example, in the scenario depicted in Figure \ref{fig5} (b), we position the pedestrian not in the center of the corridor but rather 0.5 meters away from the wall. Since our method is based on RGB images, it can accurately detect the pedestrian (Figure \ref{fig5} (c)). In this scenario, the depth values for the person and the adjacent wall are very close (as indicated by the green box in Figure \ref{fig5} (d)). As shown in Figure \ref{fig5} (e), the green box in the U-depth map corresponds to the wall and the pedestrian, and it is hard to distinguish them, potentially leading to a failure in detection. For further evaluation, we place the pedestrian in various positions, and for each frame, we perform detection using the U-depth map and our method. Subsequently, we calculate the detection success rate, and the results are illustrated in Figure \ref{fig6}.


\subsection{Autonomous obstacle avoidance}

This section evaluates the effectiveness of our planning algorithm in dynamic obstacle avoidance in the corridor simulation environment described in Section 4.1. Similarly, we position a pedestrian in the center of the corridor, who moves back and forth at a constant velocity. The depth sensing range of the UAV is 6 m. In the initial state, the UAV has no depth values for the pedestrian, and consequently, the local occupancy map used for trajectory planning does not contain this information. When the UAV detects the pedestrian, it rapidly replans the trajectory to avoid a head-on collision with the pedestrian. At the start of the experiments, the pedestrian's velocity is set to 0 and increased by 0.1 m/s until a collision occurs. The maximum speed of the UAV is set to 1.5 m/s, and the maximum acceleration is set to 3 m/s2. EGO-Planner \cite{zhou2020ego22} is also employed for comparison, and the results are presented in Table \ref{tab2}. When the pedestrian's velocity exceeds 0.8 m/s, the trajectory planned by EGO-Planner results in a collision with the person. This is primarily because it does not consider that the pedestrian is moving towards it. In contrast, our method considers the short-term future movements of the UAV, resulting in a trajectory that causes the UAV to turn in advance to avoid the pedestrian. Typically, a pedestrian’s walking velocity is around 1 m/s, indicating that our method is applicable to everyday environments.

\begin{table}
	\captionsetup{skip=3pt}
	\centering
	\caption{The maximum velocity of pedestrians without collisions.}
	\label{tab2}
	\begin{tabular}{@{}cc@{}}
		\toprule
		Method & Maximum velocity of the pedestrian\\
		\midrule
		EGO-Planner	 \cite{zhou2020ego22} & 0.8 m/s \\
		Our method & 1.7 m/s \\
		\bottomrule
	\end{tabular}
\end{table}

\begin{table}
	\captionsetup{skip=3pt}
	\centering
	\caption{The maximum velocity of pedestrians without collisions.}
	\label{tab3}
	\begin{tabular}{@{}cccc@{}}
		\toprule
		World&	Event&	EGO-Planner &	Ours \\
		\midrule
		\multirow{2}{*}{\makecell{Scene \uppercase\expandafter{\romannumeral 1} \\ (1 pedestrian)}}&	Collision &	13&	0\\
		&Freezing &	0&	0\\
		&Success &	2&	15\\
		\midrule
		\multirow{2}{*}{\makecell{Scene \uppercase\expandafter{\romannumeral 2} \\ (4 pedestrians)}} & Collision &	15&	3\\
		&Freezing &	0&	0\\
		&Success &	0&	12\\
		\bottomrule
	\end{tabular}
\end{table}

Following \cite{chen2023risk33}, we conduct tests in additional scenes. We set up two different scenes, where Scene \uppercase\expandafter{\romannumeral 1} includes 1 pedestrian, and Scene \uppercase\expandafter{\romannumeral 2} includes 4 pedestrians, and then let the UAV fly multiple times in different scenes. Each method is tested 15 times in each scene, and the number of times different events occur is counted, as shown in Table \ref{tab3}. "Collision" represents a collision occurrence, "Freezing" indicates situations where the UAV cannot plan a correct trajectory and remains stationary, and "Success" denotes normal and safe flight. It is evident that our method exhibits a significantly higher success rate. In the scene with only 1 pedestrian, no collisions occur, and in cases with multiple pedestrians, the probability of collision is low. This is due to the consideration of the future motion state of dynamic obstacles in the planning process.

\begin{figure}
	\centering
	\includegraphics[scale=0.14]{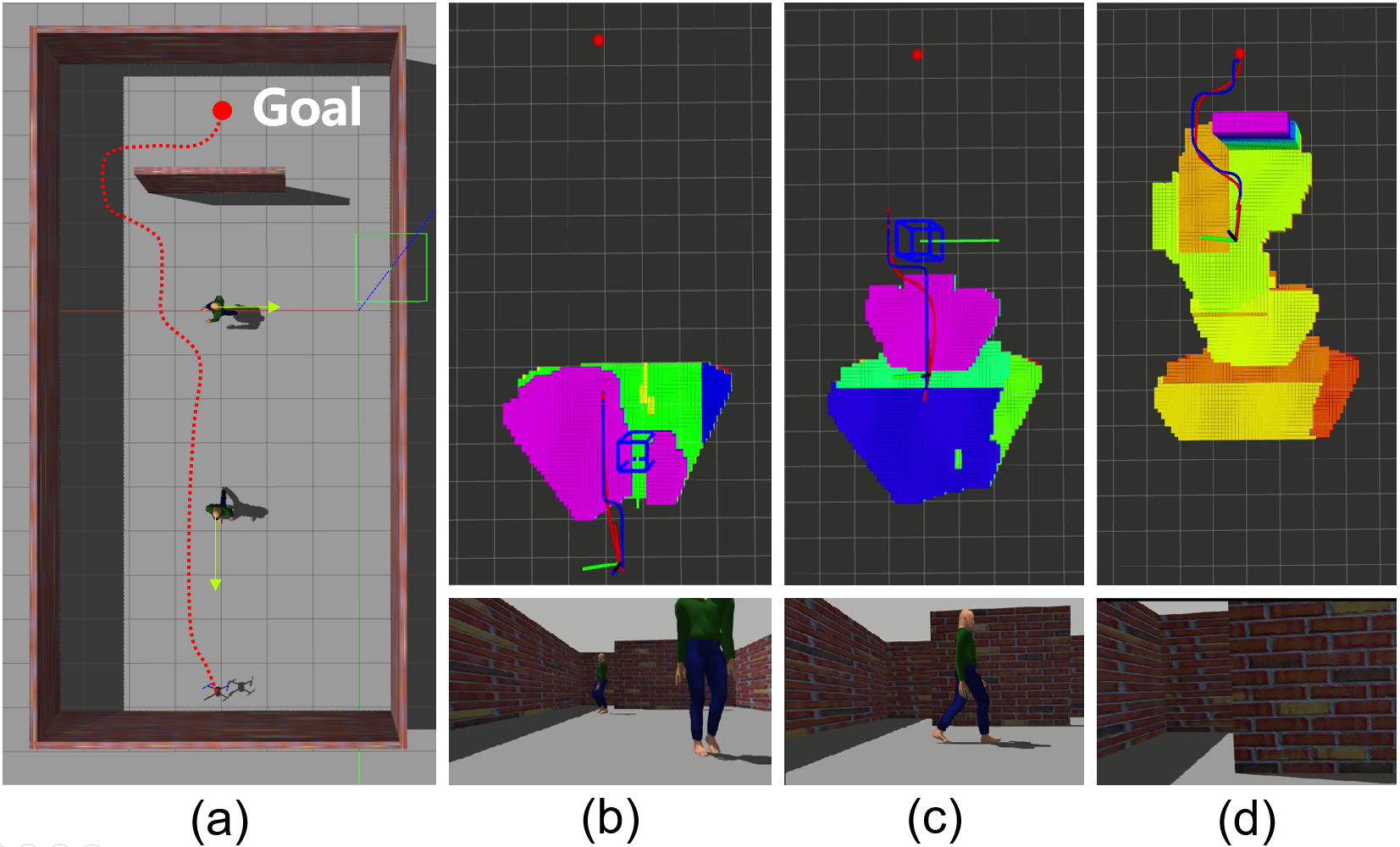}
	\caption{The planning process in the simulation environment. (a) is an overview of the scene, with the red line representing the approximate flight path of the UAV, and yellow arrows indicating the pedestrians’ movement direction as the UAV approaches. (b)-(d) represent different stages of flight, with the blue line indicating the initial trajectory and the red line showing the final trajectory.}
	\label{fig8}
\end{figure}

\begin{figure}
	\centering
	\includegraphics[scale=0.25]{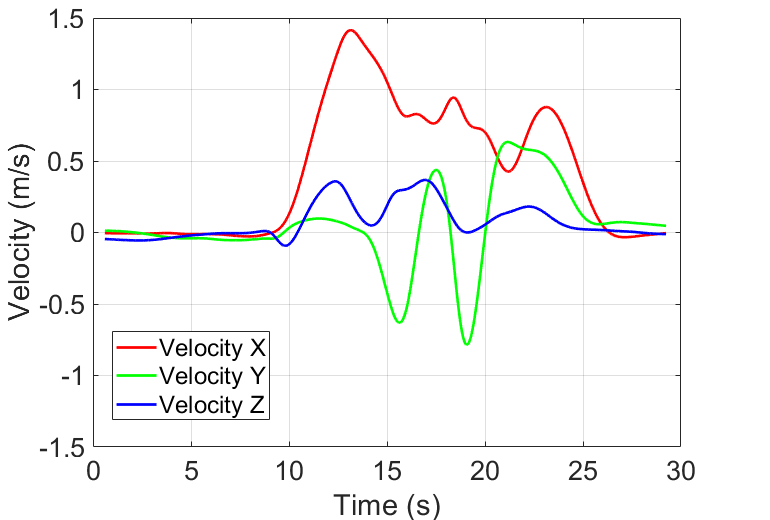}
	\caption{The velocity profile of the UAV in flight.}
	\label{fig9}
\end{figure}

A UAV planning process in a simulated environment with multiple moving objects is present in Figure \ref{fig8}. As shown in Figure \ref{fig8} (a), the goal is located behind a wall in front of the UAV, and there are two pedestrians with different movement directions performing back-and-forth motion along the path. When encountering the first pedestrian (Figure \ref{fig8} (b)), the UAV can proactively adjust its course from a distance to facilitate avoidance. When facing the second pedestrian (Figure \ref{fig8} (c)), the UAV chooses to maneuver around from behind. Overall, the UAV can effectively plan safe and flight-friendly trajectories in the presence of both dynamic and static obstacles. The velocity profile of the UAV during this process is illustrated in Figure \ref{fig9}, which aligns with the experimental requirements.

\subsection{Real-world experiments}

\begin{figure}
	\centering
	\includegraphics[scale=0.1]{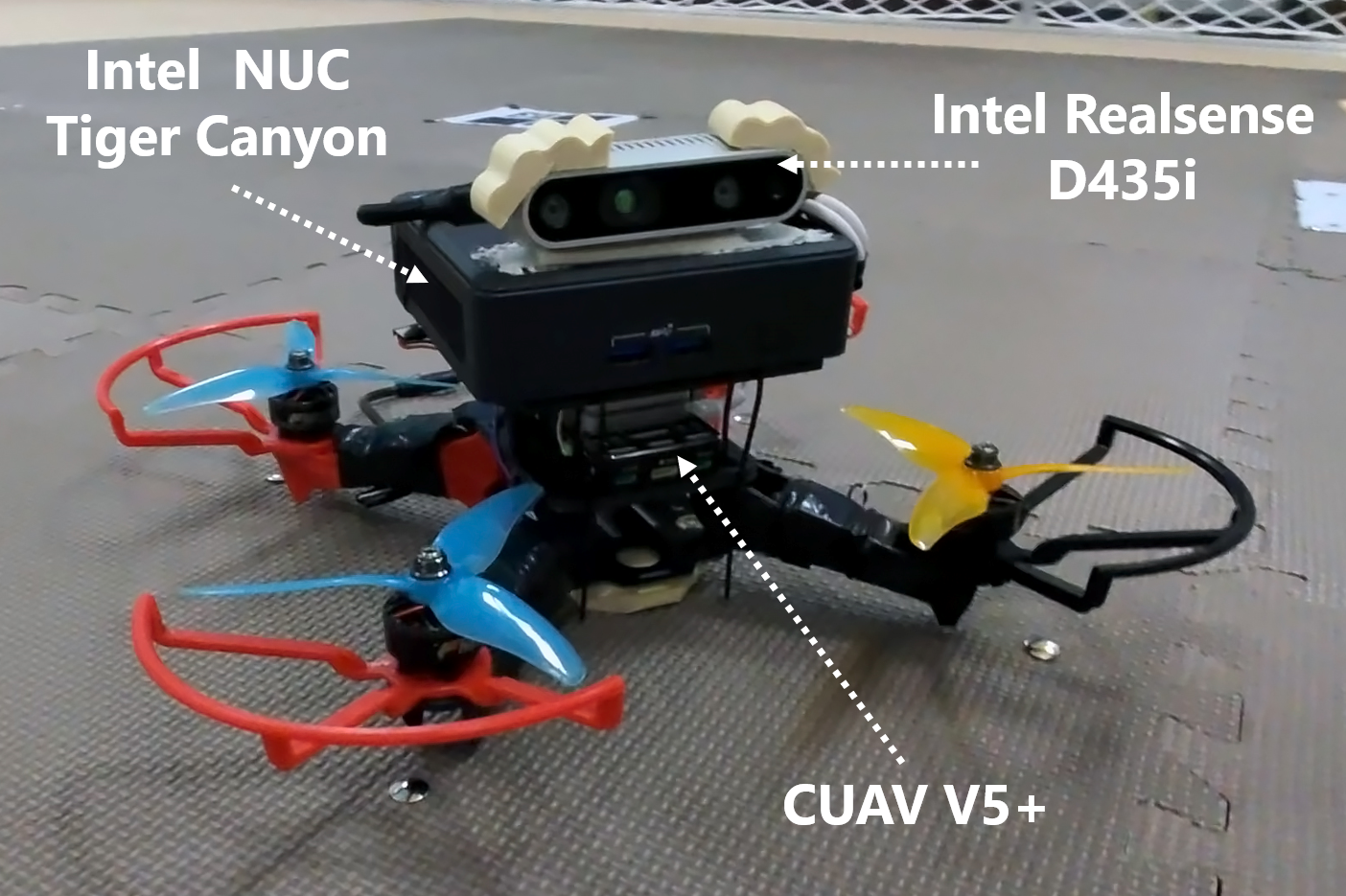}
	\caption{The hardware structure of our quadrotor.}
	\label{fig10}
\end{figure}

In real-world experiments, we employ a compact quadrotor with the hardware configuration shown in Figure \ref{fig10}. The sensor employed for depth map capture is the Realsense D435i, and the flight control system used is the CUAV V5+. The onboard computer used is a NUC equipped with an i5-1145G7 processor. This UAV utilizes the stereo infrared images captured by D435i for running VINS-Fusion \cite{qin2019general52}, thereby achieving visual localization, pose estimation, and 3D reconstruction.

\begin{figure}
	\centering
	\includegraphics[scale=0.17]{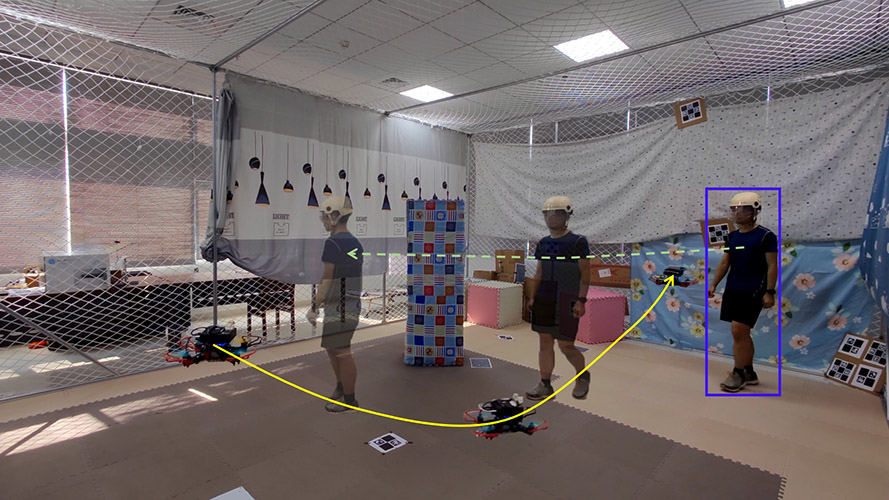}
	\caption{The testing scenario in the real world.}
	\label{fig11}
\end{figure}

We perform tests in an indoor environment, as shown in Figure \ref{fig11}. In this scenario, the pedestrian is considered as a dynamic obstacle. Initially, the drone approaches the pedestrian, and ultimately, it needs to fly to a location near the pedestrian's starting point. When planning with the EGO-Planner, there is a higher risk of colliding with the oncoming pedestrian, and mapping errors could lead to incorrect judgments. In contrast, when using our system, the UAV will go around in advance, remove the pedestrian from the local map, and finally reach the target point successfully. We also evaluated the efficiency of the system, finding that it takes approximately 200 milliseconds for the system to complete one trajectory planning cycle, which is sufficient to handle common indoor dynamic obstacles.

\section{Autonomous UAVs with LLMs}

In this section, we examine the integration of LLMs with autonomous UAVs, primarily aimed at enhancing user-drone interaction. This integration allows users to assign tasks to UAVs directly using natural language, without requiring in-depth knowledge of programming languages, while enabling UAVs to provide feedback to users in natural language or other modalities. Previous research \cite{vemprala2023chatgpt46} has demonstrated that ChatGPT \cite{chatgpt59} can take high-level textual feedback about generated code or its performance and map it to the required low-level code changes, making it accessible to users who may not possess technical expertise. For instance, ChatGPT is capable of generating complex code structures for drone navigation by utilizing only the fundamental Application Programming Interfaces (APIs) provided in the prompt, and there are many explorations about the formulation of initial prompts for various problem-solving tasks. Furthermore, GPT-4 supports multi-modal capabilities \cite{chatgpt63}, which enhance the diversity of human-machine interactions. Therefore, in this study, we use ChatGPT as an example to explore the integration of autonomous UAVs with LLMs. 

\begin{figure}
	\centering
	\includegraphics[scale=0.0303]{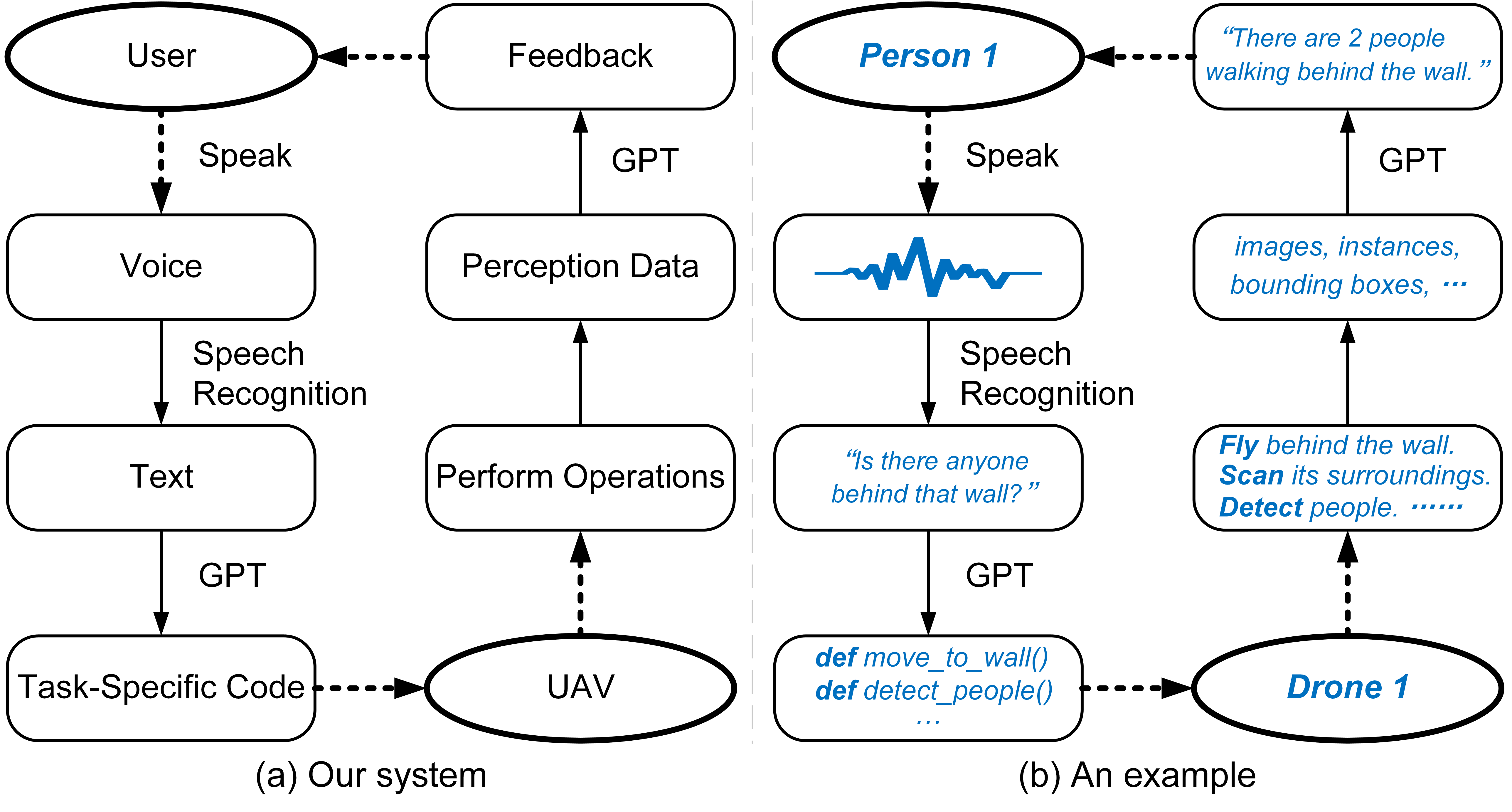}
	\caption{(a) A system combining autonomous UAVs and LLMs. (b) An example of a task executed by this system. "GPT" stands for Generative Pre-trained Transformer, representing models such as ChatGPT.}
	\label{fig12}
\end{figure}

Building upon ChatGPT and existing relevant research, we exemplify a system integrating autonomous UAVs with LLMs, as demonstrated in Figure \ref{fig12} (a). This system is based on UAVs with autonomous flight capability, and utilizes LLMs for facilitating communication between users and UAVs. Specifically, the user provides a task description to the UAV through natural language. Initially, the UAV uses voice recognition technology (such as Whisper \cite{radford2023robust76}) to convert the task description into textual information. Subsequently, it utilizes a Generative Pre-trained Transformer (GPT) (such as ChatGPT) to comprehend and analyze the task, generating the necessary code for task execution, which is then transmitted to the UAV. The UAV finally executes the programs to complete the task, all the while returning acquired data to provide feedback to the user. An exemplary task executed using this system is shown in Figure \ref{fig12} (b). A user inquires if there are persons behind a wall. Subsequently, the UAV comprehends the task, generates the requisite code, and proceeds to fly to the wall. It then scans the surroundings, detects persons, and finally expresses the collected information to the user using GPT. During this process, it can also provide real-time updates to the user, such as reporting the number of people observed during flight or its need to navigate from the right side due to obstruction by a box on the left. The simulation experiment in Figure 13 showcases an autonomous UAV integrated with LLM.

\begin{figure}
	\centering
	\includegraphics[scale=0.167]{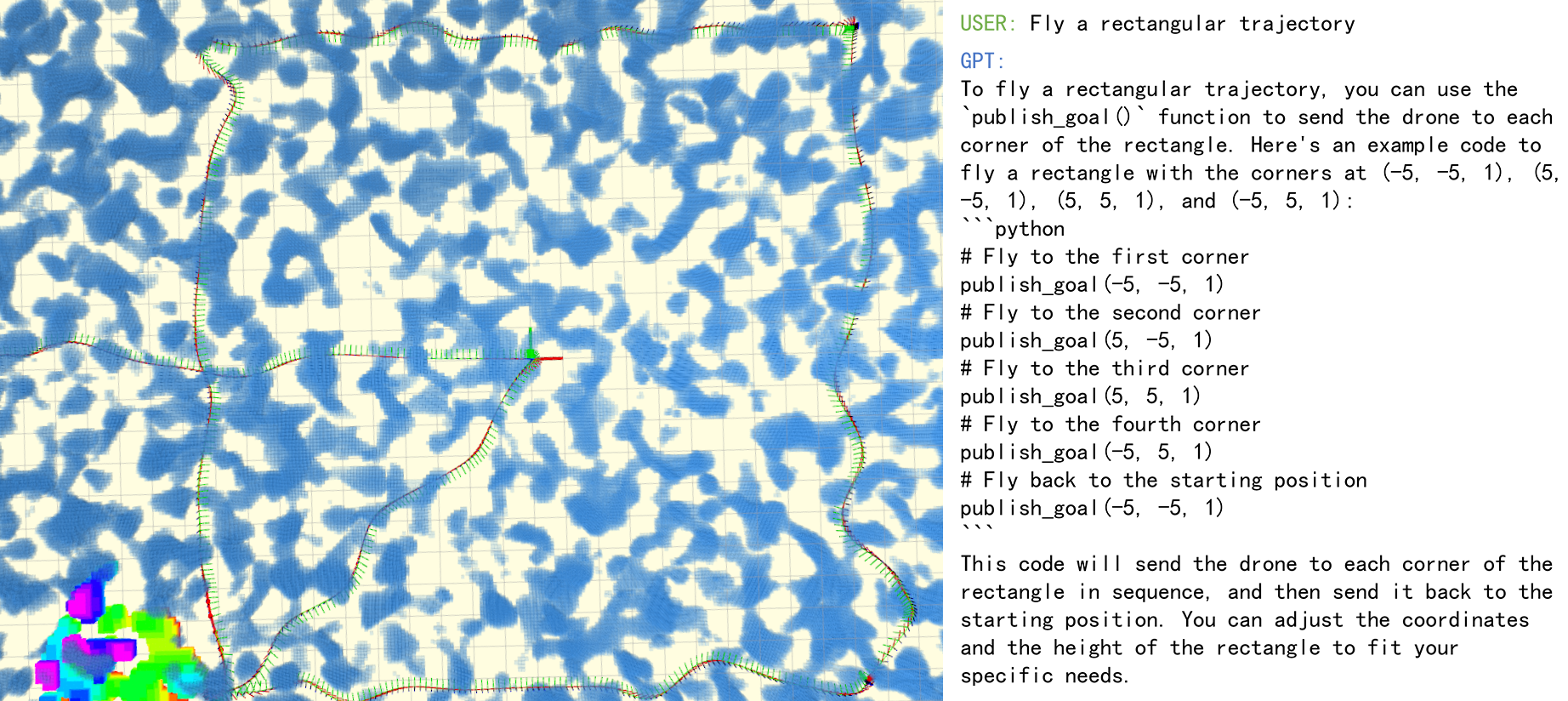}
	\caption{An example experiment of the integration of autonomous UAV with LLM in a simulation environment. The UAV, guided by an GPT-enhanced Ego-Planner, can successfully navigate a user-specified rectangular trajectory while adeptly avoiding obstacles. This experiment underscores the advanced spatial awareness and command understanding capabilities obtained by the integrating LLMs with UAV control algorithms.}
	\label{fig13}
\end{figure}

The two central components of this system are GPT and autonomous UAVs. A primary role of GPT is to generate task-specific code. The basic drone control operations (including takeoff, landing, steering, detection, and recognition) use universal code that does not need to be regenerated. Therefore, for finer drone control, it is advantageous to establish a library of APIs for robot control and perception, which can be readily adapted for use with GPT. Based on previously defined APIs, GPT can build a new function for more complex tasks \cite{vemprala2023chatgpt46}. And given the functions, GPT can generate controls for long-step tasks \cite{wake2023chatgpt77}. As for autonomous UAVs, they must possess the capability for safe execution of diverse operations and fundamental decision-making abilities. For the instance in Figure \ref{fig12}(b), when the UAV is required to fly to the other side of a wall, it should at least be able to perform trajectory planning in a static environment. Given the potential presence of moving objects, such as people, it should also be capable of avoiding dynamic obstacles. And given the task specified by the user that involves more multi-steps, it should be able to break down the task and generate a schedule, then execute sequentially. In our proposed approach, significant interaction occurs between the UAV and real-time data, enabling dynamic obstacle avoidance—an important aspect of drone control. It is worth noting that while LLMs inherently face challenges in directly leveraging real-time sensor data for immediate responses, their integration with our system plays a different but critical role. In this case, the LLM acts as a bridge, providing a more user-friendly interface that effectively exploits the capabilities of real-time drone methods. By combining the advanced technical capabilities of drones in dynamic obstacle navigation with the intuitive, user-centered interface provided by LLM, our approach makes a step towards realizing intelligent UAVs. This synergy not only improves operational efficiency, but also significantly improves ease of use, making advanced UAV technology more accessible to a wider user base.

\section{Conclusion and future work}

This paper presents an autonomous planning system for quadcopter UAVs suitable for dynamic environments. We employ a lightweight neural network to detect dynamic obstacles, then use KF for object tracking and prediction, and take into account both static environments and dynamic obstacles in trajectory planning. For trajectory generation, we utilize a B-spline-based trajectory search algorithm and optimize the trajectory with multiple constraints, resulting in trajectories that are not only safer but also better compliant with the UAV’s kinematics. Experimental results in both simulation and real-world environments demonstrate that our method can successfully detect and avoid collisions with obstacles in dynamic environments in real-time. Building upon this, we explore the integration of UAVs equipped with autonomous flight systems with LLMs to serve upper-level applications. This represents an important avenue for future development and an area of focus for our forthcoming research efforts.

\section*{Acknowledgments}

Special thanks go to the indoor UAV testing experimental site of the Center for Gravitational Experiments of Huazhong University of Science and Technology and Zipeng Jiang for their assistance. This study was funded by the NSFC (41901407), the enterprise-funded research project titled "Research on Key Technologies of Intelligent Multi-Rotor UAVs" (k23-4201-026), and the College Innovation and Entrepreneurship Training Program. 

{\small
\bibliographystyle{ieee_fullname}
\bibliography{egbib}
}

\end{document}